\documentclass[11pt,a4paper]{article}
\usepackage{authblk}
\usepackage[hyperref]{emnlp2020}
\usepackage{times}
\usepackage{latexsym}

\usepackage{microtype}

\usepackage{amsmath}
\usepackage{amssymb}
\usepackage{times}
\usepackage{latexsym}
\usepackage{graphicx}
\usepackage{url}
\usepackage{subcaption}
\usepackage{multirow}
\usepackage{mathtools}
\usepackage{color, colortbl}
\usepackage{paralist}
\definecolor{Gray}{gray}{0.9}
\definecolor{LightCyan}{rgb}{0.88,1,1}
\definecolor{Pink}{rgb}{1,0.98,1}

\aclfinalcopy 
\def\aclpaperid{561} 

\newcommand{\tvt}{2\ vs.\ 2 }
\newcommand{\jab}{\textit{Jabberwocky} }
\newcommand{\sen}{\textit{Sentence} }
\newcommand{\wl}{\textit{Word-list} }

\usepackage{xr}

\makeatletter
\newcommand*{\addFileDependency}[1]{
  \typeout{(#1)}
  \@addtofilelist{#1}
  \IfFileExists{#1}{}{\typeout{No file #1.}}
}
\makeatother

\newcommand*{\myexternaldocument}[1]{
    \externaldocument{#1}
    \addFileDependency{#1.tex}
    \addFileDependency{#1.aux}
}
\myexternaldocument{emnlp2020_appendix}

\title{From Language to Language-ish: How Brain-Like is an LSTM's Representation of Nonsensical Language Stimuli?}

\author[1,2, *]{Maryam Hashemzadeh}
\author[3,4]{Greta Kaufeld}
\author[1,2]{Martha White}
\author[3,5]{\\ Andrea E. Martin}
\author[1,2,6]{Alona Fyshe}
\affil[1]{\small{ Department of  Computing Science, University of Alberta, Alberta, Canada} }
\affil[2]{\small{Alberta Machine Intelligence Institute (Amii), Alberta, Canada}}
\affil[3]{\small{Max Planck Institute for Psycholinguistics, Nijmegen, The Netherlands}}
\affil[4]{\small{International Max Planck Research School for Language Sciences, Nijmegen, The Netherlands}}
\affil[5]{\small{Donders Institute for Brain, Cognition, and Behaviour, Radboud University, Nijmegen, The Netherlands}}
\affil[6]{\small{Department of Psychology, University of Alberta, Alberta, Canada}}
\affil[*]{\small{\textit {Corresponding author, hashemza@ualberta.ca}}}
\affil[ ]{\small{\textit {\{whitem,alona\}@ualberta.ca, \{greta.kaufeld,  andrea.martin\}@mpi.nl}}}

\date{} 

\begin{document}
\maketitle
\begin{abstract}
The representations generated by many models of language (word embeddings, recurrent neural networks and transformers) correlate to brain activity recorded while people read.  However, these decoding results are usually based on the brain's reaction to syntactically and semantically sound language stimuli. In this study, we asked: how does an LSTM (long short term memory) language model, trained (by and large) on semantically and syntactically intact language, represent a language sample with degraded semantic or syntactic information?  Does the LSTM representation still resemble the brain's reaction?  We found that, even for some kinds of nonsensical language, there is a statistically significant relationship between the brain's activity and the representations of an LSTM. This indicates that, at least in some instances, LSTMs and the human brain handle nonsensical data similarly. 
\end{abstract}

\section{Introduction}
When people read or listen to language, brain imaging studies have shown us that the brain's activity correlates to LSTM (long short term memory) state representations for the same text~\citep{jain2018incorporating, toneva2019interpreting}.  In those studies (and others like them) the stimuli used to test for this correlation was based on language with no errors.\footnote{Nonsensical language is often used when measuring Event Related Potentials.  Here we speak of decoding studies only.}  This implies that during the processing of within-distribution data (i.e. well-formed sentences/stories), LSTMs and the human brain show similar representational patterns.  But what happens when language is out-of-distribution (e.g. nonsensical sentences or pseudo-words)?  Can we expect that an LSTM will still compute contextual states in a way that resembles how the human brain reacts?   I.e. is there a correlation between LSTM representations and neural activity when the stimuli is not a predictable language sample? Answering these questions could provide evidence that an LSTM is able to generalize to new data in a human-like way, even when the new data is unlike anything it encountered during training.   Our answers could also help psycholinguists reason about the efficacy of nonsensical sentences and pseudo-words as syntax-only stimuli controls.

Here we use brain imaging data (Electroencephalography, EEG) collected in three conditions, \textit{Sentence}: well-formed grammatical sentences, \textit{Jabberwocky}: pseudo-word sentences that preserved word order, morphosyntax, and sentential prosody without lexical or compositional semantics, and \textit{Word-list}: the words of the \sen condition in a pseudo-random order without sentence prosody, syntax, or compositional meaning.  We ran a character-level LSTM model on the stimuli, and trained a \emph{decoding model} to predict the LSTM's internal representations from EEG signals.  Using data from the \sen condition, we corroborated previous results and showed that LSTM representations are correlated with brain activity for within-distribution language.  But, when it came to nonsensical language stimuli, it was unclear if LSTM representations would still correlate to brain activity.  Our original hypothesis was that LSTM representations for out-of-distribution language  would no longer correlate to brain activity.  However, we found that our decoding model worked quite well even when all content words of the stimuli were pseudo-words (\textit{Jabberwocky}). 

To summarize, we show that:

\begin{compactitem}
\item  Our decoding models work well in both the \sen and \jab conditions, but not in the \wl condition.
\item  The syntactic signatures available in \sen and \jab LSTM representations are similar, and can be predicted from either the \sen or \jab EEG.
\item For some LSTM representations, the decoding model's \emph{weight maps} generalize between \jab and \sen EEG data.  
\item From our results, we can infer which LSTM representations encode semantic and/or syntactic information.  We confirm using syntactic and semantic probing tasks. 
\end{compactitem}
Our results show that there are similarities between the way the brain and an LSTM represent stimuli from both the \sen (within-distribution) and \jab (out-of-distribution) conditions.

\section{Materials and Methods}

\subsection{Data description}

Our data was originally collected to contrast the brain's response to language samples that vary the amount of semantic and syntactic information~\cite{kaufeld2020linguistic}. 
The dataset consists of EEG recordings ($64$ channels, $500$ Hz sampling rate) of $27$ native Dutch speakers ($9$ males; mean age$=23$). The participants listened to a native Dutch speaker in three conditions: \textit{Sentence}, \textit{Jabberwocky}, and \textit{Word-list}. Each condition has 80 sentences, and all \sen and \jab stimuli sentences share the same grammatical structure. 
 
The \textit{Sentence} stimuli contain two coordinate clauses and a conjunction with the structure \textit{[Adj N V N Conj Det Adj N V N]}, and contain lexical semantics, compositional semantics, and syntax.
\textit{Word-list} consists of the same ten words as \textit{Sentence} but in a pseudo-random order with infeasible syntactic structures (either \textit{[V V Adj Adj Det Conj N N N N]}, or \textit{[N N N N Det Conj V V Adj Adj]}).  The \wl condition leaves orthography/phonology intact and contains lexical semantics, but not compositional semantics or syntax. 
For \textit{Jabberwocky}, words from the \sen condition are replaced with pseudo-words created with the Wuggy generator~\cite{keuleers2010wuggy}.  Crucially, the {\bf \jab pseudo-words appear in the same order as the corresponding words in the \sen condition}. The Wuggy generator alters words in a way that obeys the phonotactic and morphosyntactic constraints of a language, but eliminates semantic meaning.  The \jab condition contains syntax (and morphosyntax, which is preserved by Wuggy).
Amongst psycholinguists and cognitive neuroscientists, it is widely accepted that Jabberwocky does not contain lexical or compositional semantics, and a Jabberwocky condition is often used to control for semantics \citep{Humphries2006,Fedorenko2012,Friederici2000}. Anecdotally, native Dutch speakers typically cannot guess the true word when presented with the pseudo-word.

Stimuli examples:
\begin{compactitem}
    \item \textit{Sentence}: Lange mannen bouwen huisjes en de lieve honden brengen planken.
    (Tall men build houses and the sweet dogs bring boards.)
    \item \textit{Jabberwocky}: Lalve wanzen botren raasjes en de reeve rorden brargen sponken. 
    \item \textit{Word-list}: planken mannen huisjes honden de en bouwen brengen lange lieve
\end{compactitem}

In the \jab condition the determiners and conjunctions are not pseudo-words.  To fairly compare the conditions, we removed these words from all three conditions during our analyses. 
Due to the nature of spoken language, the time-duration each of word/pseudo-word differs.  To account for this, we considered the first 400 ms of EEG after word/pseudo-word onset. 

To improve the EEG's signal to noise ratio, we average the EEG recording for a given sentence across all subjects. Though this reduces participant-specific signal, we have found it to be the best way to decode from EEG data. For this data, models trained on only one subject did not perform above chance.
For each word of each stimulus sentence $S$, we concatenated the recording from every electrode into one vector $R_t \in \mathbb{R}^{1 \times D}$ where $D$ is the total number of readings across all sensors (here $D=12800$: $64$ sensors $\times\  200$ time points).

\subsection{Decoding model}

\begin{figure*}
\centering
\includegraphics[width=\textwidth]{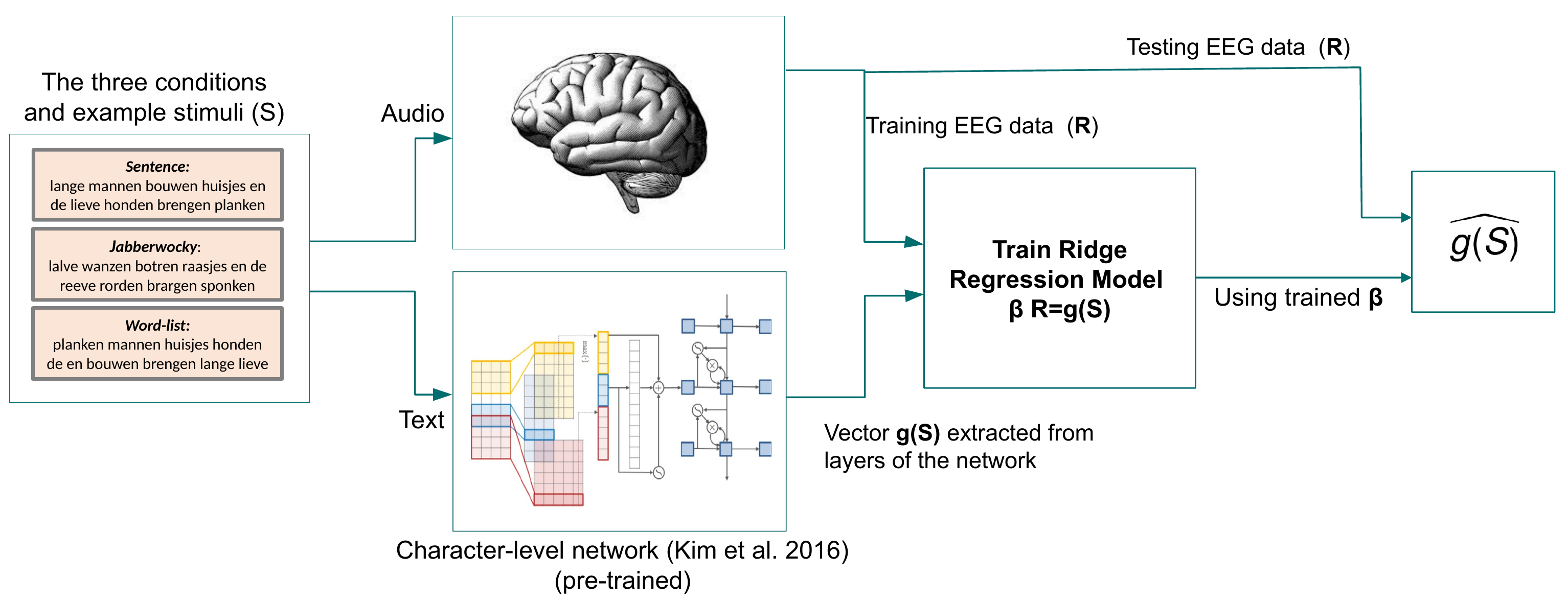}
\caption{Decoding model. Each stimulus sentence is fed to a pre-trained language model to create a non-linear context-based representation. The hidden representations for a sentence ($S$) are extracted from each layer $g(S)$. Our ridge regression model is trained to use the EEG signal $R$ to predict $g(S)$.
}
\label{fig:schematic}
\end{figure*}

The aim of a decoding model is to find a mapping function $f(R_t) \rightarrow g(S_{1:t})$ between an EEG recording $R_t$ of the brain's response to word $w_t$ and a language model's representation of stimulus $S_{1:t}$ (the words of a sentence up to and including word $w_t$). 
Our methodology closely followed~\cite{jain2018incorporating}. We instantiate our mapping function in two steps: 
\begin{compactenum}
    \item $g(S_{1:t}) \in \mathbb{R}^{1 \times P} $: an LSTM's $P$-dimensional representation for word $w_t$, conditioned on context $w_{1}, \ldots w_{t-1}$.
    \item $f(R_t) $: a regularized linear regression to map the EEG signal $R_t$ to $g(S_{1:t})$.
\end{compactenum}
Figure~\ref{fig:schematic} shows a schematic of the decoding model.

\paragraph{1- Derive LSTM representations ($g(S)$):}

The \textit{Jabberwocky} stimuli are made of pseudo-words, so we needed a language model that can handle out-of-vocabulary input. We used the state-of-the-art character-level LSTM language model proposed by \citet{kim2016character}, but used three LSTM layers based on previous decoding work \cite{jain2018incorporating}. 
This LSTM operates on the characters of incoming words (so it can handle pseudo-words), but it produces predictions at the word level. Each input character has its own embedding, which are concatenated and fed to convolutional layers. The convolved values are passed to a highway network, whose output is fed through three stacked LSTM layers before predicting the next word.
In the decoding analyses that follow, the $g(S)$ vectors we analyze are (1) the concatenation of the character embeddings called \textit{Embedding} layer, (2) the concatenation of the Convolutional layers called \textit{Conv}, and (\{3-5\}) the three LSTM layers called \textit{LSTM1-3}.  We will use the term \emph{LSTM} to refer to the full character-based model, \emph{LSTM representation} to refer to any of the $g(S)$ vectors types, and \emph{LSTM layer} or \emph{LSTM1-3} to refer specifically to the LSTM layers within the larger LSTM model.

We trained the LSTM on one million sentences from Dutch Wikipedia. We set the number of epochs to $40$, batch size is 50, and sequence length is $20$. We used a stochastic gradient descent optimizer with sparse categorical cross-entropy loss. The initial learning rate is  $0.8$ with inverse time decay rate $0.5$.

For Dutch Wikipedia, the average test perplexity of our model is $108.12$.
When the inputs are the \textit{Sentence} stimuli, the average perplexity is higher: $317.91$.  This is likely because the coordinate clauses within each stimulus are only 4 words long, which reduces the effective context.
When the inputs are the \textit{Jabberwocky} pseudo-words and the outputs are the corresponding \textit{Sentence} next word, the perplexity is $325.12$.
These \sen and \jab perplexities are not significantly different ($p= 0.967$). We calculated the average perplexity on the \textit{Word-list} stimuli to be $1008.23$, which indicates that (as expected) the network cannot predict the next words in the \textit{Word-list} stimuli. This also shows that while the \jab and \sen perplexities are higher than on Wikipedia, they are much lower than for stimuli with no contextual information.  

For comparison, we also experimented with non-contextual word embeddings from \citet{grave2018learning}. This 300-dimensional model is pre-trained on Dutch Wikipedia using Continuous Bag of Words (CBOW) with position-weights.

\paragraph{2- Regularized linear regression ($f(R)$):}
We used ridge regression to test if the EEG data correlates with the  word/pseudo-word representations. The regression function $f(R_t)$ is a linear transformation of $R_t$ to predict the $P$-dimensional $g(S_{1:t})$: $f(R_t) = R_t \beta$ where $\beta \in \mathbb{R}^{D \times P}$.

\subsection{Measuring model accuracy}

We used Monte Carlo (MC) cross-validation to evaluate our decoding models. MC cross-validation affords a more stable estimate of model accuracy, and allows for statistically-sound comparisons of model performance. During each of our 200 MC samples, we swept the regression regularization parameter among the values in range $[0.1, 200]$ using $5$-fold cross-validation on the training data only.  

We use a 2 vs. 2 classification test to assess the performance of the learned model \cite{mitchell2008predicting,Fyshe2019}. 
During each cross-validation trial we randomly create groups of two from the held-out samples. Using a model fit to the training data, we produce predicted representations for the held-out samples.  For simplicity, let $y^{i}_t=g(S^{i}_{1:t})$ be the contextual representation for word $w_t$ of sentence $i$.  
Then, for each group of 2 test samples ($S^{i}_{1:t1},S^{j}_{1:t2}$), we perform a \emph{\tvt test} using the true representations $(y^{\ i}_{t1},y^{\ j}_{t2})$ and predicted representations  $(\widehat{y}^{\ i}_{t1},\widehat{y}^{\ j}_{t2})$.  The \tvt test compares the sum of cosine similarity for correctly matched the true and predicted vectors:
\setlength{\abovedisplayskip}{2pt}
\setlength{\belowdisplayskip}{2pt}
\begin{align}
cos({y}^{i}_{t1},\widehat{y}^{\ i}_{t1}) + cos({y}^{\ j}_{t2},\widehat{y}^{\ j}_{t2}), \label{e.match}
\end{align}
to the sum of cosine similarity of the mismatched vectors:
\begin{align}
    cos({y}^{\ i}_{t1},\widehat{y}^{\ j}_{t2}) + cos({y}^{\ j}_{t2},\widehat{y}^{\ i}_{t1}). \label{e.mismatch}
\end{align}
If Eq~\ref{e.match} is greater than Eq~\ref{e.mismatch}, the \tvt test passes. \tvt accuracy is the percentage of correct \tvt tests, and chance \tvt accuracy is $0.5$.  In addition to \tvt accuracy, we also report mean-squared-error of the learned model in Appendix~\ref{app:B}.

To test for statistical significance, we used permutation tests. The LSTM representations for the stimuli were randomly shuffled such that the true representations $g(S_t)$ were no longer correctly matched to the EEG data.  We then trained and tested our decoding models as described above using  $>1000$ random permutations.  These results represent the expected distribution of \tvt accuracy when there is no relationship between the EEG data and the LSTM representations.
From that (null) distribution we can compute $p$-values for our observed accuracy on the un-permuted representations. 
We correct for multiple comparisons using the Benjamini-Hochberg-Yekutieli False Discovery Rate (FDR) procedure \cite{benjamini2001control} using $\alpha=0.05$.

For our models to perform above chance, there must be correlates of particular aspects of language (such as semantics or syntax) present in the brain activation data $R$, and in the corresponding contextual representation ($g(S)$).  Furthermore, our decoding model assumes a linear relationship between $R$ and $g(S)$. If our models do not perform above chance, any of the above conditions may be violated; our analyses are not designed to differentiate between the failure cases.




\section{Results}

We were interested in comparing the representations generated by an LSTM to that of the human brain, in response to both within- and out-of-distribution language.  Our \textit{Sentence} stimuli, which represent within-distribution language, contain semantic and syntactic information.  We used two kinds of out-of-distribution stimuli: \textit{Jabberwocky}, which was designed to have syntactic information only, and \textit{Word-list}, which has only semantic information.  We attempted to learn a mapping from EEG to LSTM representations (to test if the LSTM and brain handle the stimuli similarly). 
To begin, we examined the difference in the semantic and syntactic information encoded by each of the LSTM representations.
Then, we developed analyses to test for a similarity in the representation of semantic and syntactic information across the experimental conditions.   
We investigated using the following questions:

\begin{compactenum}
\item Is there a difference in the semantic/syntactic information captured by the LSTM representations? (Probing tasks) 
\item Can we learn a mapping from the EEG data to the LSTM representations in the \textit{Sentences, Jabberwocky}, or \textit{Word-list} conditions? Is there a difference in performance across the different LSTM representations? (Analysis 1: test for semantic and/or syntactic information)
\item If there is syntactic information present in the \textit{Sentences} and \textit{Jabberwocky} LSTM representations, is it exchangeable?  (Analysis 2: swap the $g(S)$ conditions)
\item Do the actual \emph{patterns} learned by the decoder generalize to EEG from the other condition? 
(Analysis 3: swap $R$ at \emph{test} time only)
\end{compactenum}
The EEG analyses are summarized in Table~\ref{t:exp1}.

\begin{table}\small
  \begin{center}
    \begin{tabular}{|c|c|c|c|c|c|} 
    \hline
      \textbf{} & \textbf{} & \textbf{Train} & \textbf{Train } & \textbf{Test } & \textbf{Test } \\
      \textbf{Analysis} & \textbf{Case} & \textbf{EEG} & \textbf{ $g(S)$} & \textbf{EEG} & \textbf{ $g(S)$} \\
      \hline
      \rowcolor{Gray}
      &\textbf{1} &\textit{Sen} & \textit{Sen} & \textit{Sen} & \textit{Sen} \\ \cline{2-6}
      \rowcolor{Gray}
     &\textbf{2} &\textit{Jab} & \textit{Jab} &\textit{Jab} & \textit{Jab} \\ \cline{2-6}
     \rowcolor{Gray}
     \multirow{-3}{*}{\textbf{1}} &\textbf{3} &\textit{WL} & \textit{WL} &\textit{WL} & \textit{WL} \\ \cline{2-6}
      \hline
      \rowcolor{LightCyan}
      &\textbf{1} &\textit{Sen} & \textit{Jab} & \textit{Sen} & \textit{Jab} \\ \cline{2-6}
      \rowcolor{LightCyan}
     \multirow{-2}{*}{\textbf{2}} &\textbf{2} &\textit{Jab} & \textit{Sen} &\textit{Jab} & \textit{Sen}  \\ \cline{2-6}
      \hline
      \rowcolor{Pink}
      &\textbf{1} &\textit{Sen} & \textit{Sen} & \textit{Jab} & \textit{Sen} \\ \cline{2-6}
      \rowcolor{Pink}
     \multirow{-2}{*}{\textbf{3}} &\textbf{2} &\textit{Jab} & \textit{Jab} &\textit{Sen} & \textit{Jab} \\ \cline{2-6}
      \hline
    \end{tabular}
    
    \caption{Data description for each analysis. Sen: Sentence, Jab: Jabberwocky, WL: Word-list. 
    Analysis 1: EEG \& $g(S)$ from the same condition. 
    Analysis 2: $g(S)$ swapped between conditions. 
    Analysis 3: EEG swapped between conditions \emph{at test time only}. 
    }\label{t:exp1}
  \end{center}
\end{table}

\subsection{Probing tasks}

\label{sec:prob}
\begin{figure}[t]
\centering
\includegraphics[width=\linewidth]{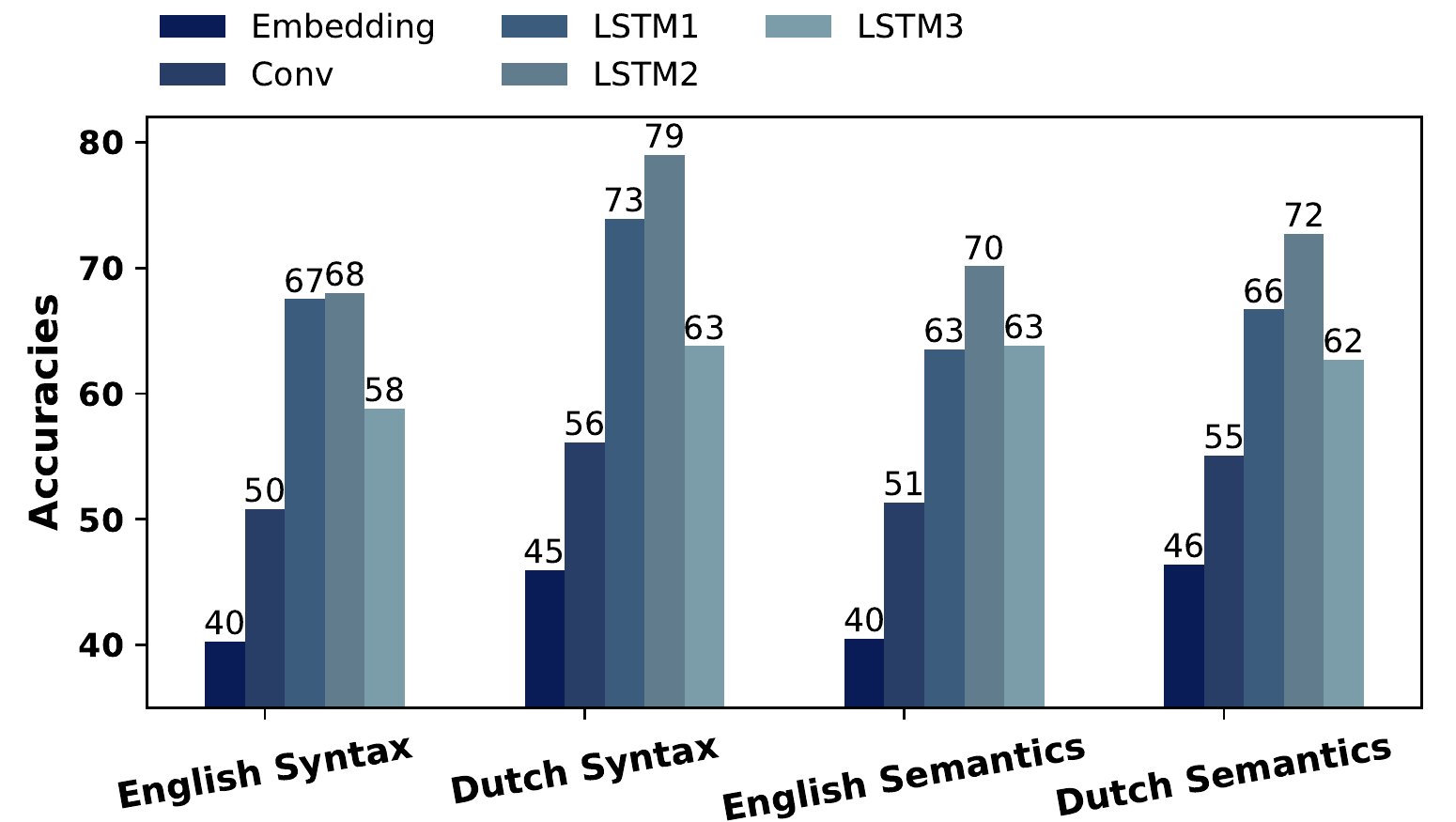}  
\caption{Average accuracies for the semantic/syntactic probing tasks using LSTM representations from Dutch or English LSTM language models.}
\label{fig:barprob}
\end{figure}

Previous work has found that LSTM layers encode differing amounts of information about semantic meaning and syntactic structure ~\cite{mccann2017learned, peters2018deep}. To investigate the behavior of our LSTM, we used several probing task benchmarks.  Because there are more available benchmarks for English, we also trained an identical LSTM architecture using English Penn Treebank (PTB)~\cite{marcus1993building}, and checked the probing task results for consistency against the Dutch results. The English semantic and syntactic probing tasks are from \citet{conneau2018you}, and the Dutch from \citet{eichler-etal-2019-linspector}. A description of each task is given in the Appendix (Table \ref{t:probDef}). 

For each probing task we trained an MLP classifier with 2 hidden layers of $100$ units. The MLP input is the average of the LSTM representations for a sentence, and the output is the predicted class of the sentence (e.g. past tense verb). Note that the sentences here are not from our stimuli, but rather from the probing tasks themselves.

The average accuracies for the English and Dutch probing tasks are shown in Fig.~\ref{fig:barprob}, and individual task accuracies appear in Table~\ref{t:resultProbEng} of Appendix~\ref{app:A}. 
We were reassured to see the performance of the English and Dutch LSTMs show similar patterns. Both the Embedding and Conv layers perform poorly on the semantic and syntactic tasks.
We see the strongest evidence for syntax in LSTM1 and LSTM2, and the strongest evidence for semantics in LSTM2.

\begin{figure}[ht]
\centering
\includegraphics[width=\linewidth]{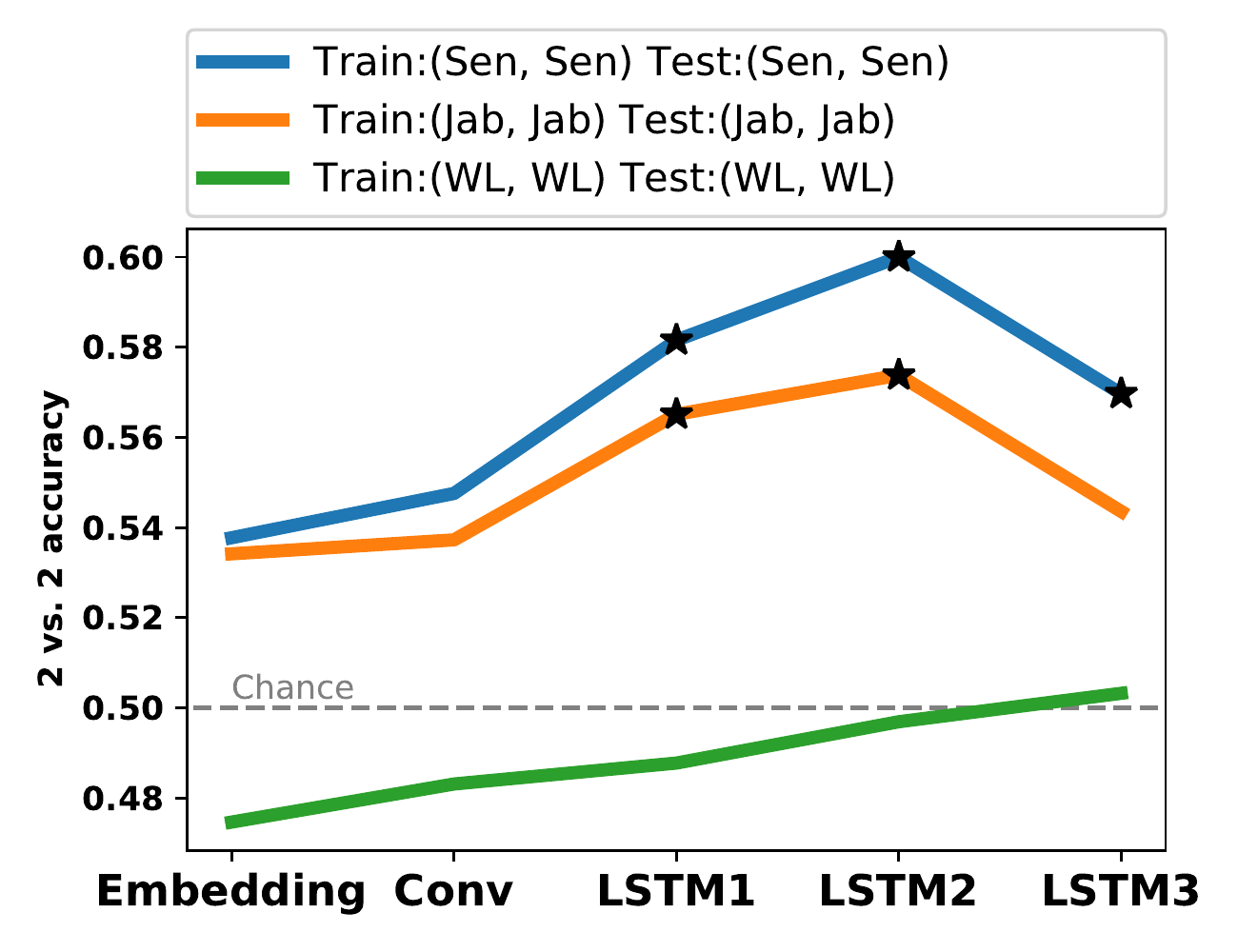}  
\caption{Analysis 1 (Test for semantic and syntactic information): \tvt accuracy with  $g(S)$/EEG from the same condition.   The $x$-axis denotes LSTM representation ($g(S)$).
Legend denotes EEG/LSTM representations used for train/test: (EEG condition, LSTM condition). ``\textit{Sen}'': \textit{Sentence}, ``\textit{Jab}'': \jab, ``\textit{WL}'': \textit{Word-list}. {\bf $\star$}: above chance ($p < 0.05$, FDR corrected).
}
\label{fig:exp1}
\end{figure}

\subsection{Test for semantic and/or syntactic information (Analysis 1)}

To test for the correlation of semantic and/or syntactic information between the EEG and LSTM representations, we measured the accuracy of a decoding model trained with data from the same condition. This is Analysis 1 from Table~\ref{t:exp1}, and results are in Fig.~\ref{fig:exp1}. 

Based on the probing results, for the \sen stimuli we expected to see highest performance for LSTM2 (contains semantic and syntactic information), and somewhat lower performance for LSTM1 (strong syntax performance, but lower semantics). 
For the \jab condition, we expected to see strongest performance for the syntactically rich LSTM1 and LSTM2.  For the \wl condition, we were unsure if the contextual representations would work at all, given that the random ordering of words removes the sentence's context.  

In the \textit{Sentences} condition, the accuracy is statistically above chance for LSTM layers 1-3 ($0.581$, $0.600$, and $0.569$ respectively, $p < 0.05$, FDR corrected).  This matched our predictions based on the probing tasks, and shows that LSTM3 has sufficient syntactic/semantic information for the decoding task.  
In the \textit{Jabberwocky} condition, only the accuracies of the LSTM1 and LSTM2 are statistically above chance with ($0.565$ and $0.573$ respectively, $p < 0.05$, FDR corrected), which again matched our predictions based on the probing tasks.  The \sen condition conveys both semantic and syntactic information, and so the decoding model produces higher accuracy than the \textit{Jabberwocky} condition, which lacks semantics. For both \jab and \sen conditions, LSTM2 shows accuracy higher than LSTM1 and LSTM3, which is consistent with previous decoding work showing that middle LSTM layers outperformed early and late layers~\cite{jain2018incorporating, toneva2019interpreting}.

When the decoding model is trained on data from the \textit{Word-list} condition, no representation performs significantly different from chance ($p > 0.05$, FDR corrected).  Because of this poor performance, Analyses 2 and 3 do not consider the \wl condition.  The accuracies for the Embedding and Conv layers are not significantly above chance for any condition ($p > 0.05$, FDR corrected).

We also trained our decoding models with  non-contextual CBOW representations, and found the \tvt accuracy to be $0.55$ for the \sen condition, and $0.54$ for the \textit{Word-list} condition, neither of which are above chance.
Since the \jab stimuli are pseudo-words, we cannot test the \tvt accuracy using this word-level model.

\subsection{Swap the $g(S)$ conditions (Analysis 2)} 
Analysis 1 showed that some LSTM representations could be decoded in the \sen and \jab conditions.  This tells us there is a relationship between the information in some LSTM representations and the corresponding EEG data.  But, the syntactic signatures that contribute to that relationship could be condition-specific.  That is, the syntactic EEG signals driven by \jab could be fundamentally different from those in the \sen condition.  

To test if the syntactic signatures in the \sen and \jab conditions are exchangeable (i.e. similar in some way), we examined the accuracy of the decoding model in two cases: 1) using the EEG signals from the \textit{Sentence} condition to predict the $g(S)$ vectors from the \textit{Jabberwocky} stimuli, and 2) using the EEG signals from the \textit{Jabberwocky} condition to predict the $g(S)$ vectors from the \textit{Sentences} stimuli (see Table \ref{t:exp1}, Analysis 2).    Because the \jab LSTM representations do not contain semantic information, this analysis will also tell us the degree to which the \textit{Sentences} EEG/LSTM results in Analysis 1 leveraged semantic information. Because it is so central to this analysis, we again note that {\bf the \jab stimuli are composed of pseudo-words derived from the \sen stimuli, and the word order is maintained.}  That is, the first word of sentence 1 in the \jab condition is a pseudo-word transformation of the first word from sentence 1 of the \sen condition.  Thus, we can interchange the corresponding representational vectors.  


\begin{figure*}
\centering
\begin{subfigure}{.48\textwidth}
  \centering
  \includegraphics[width=\linewidth]{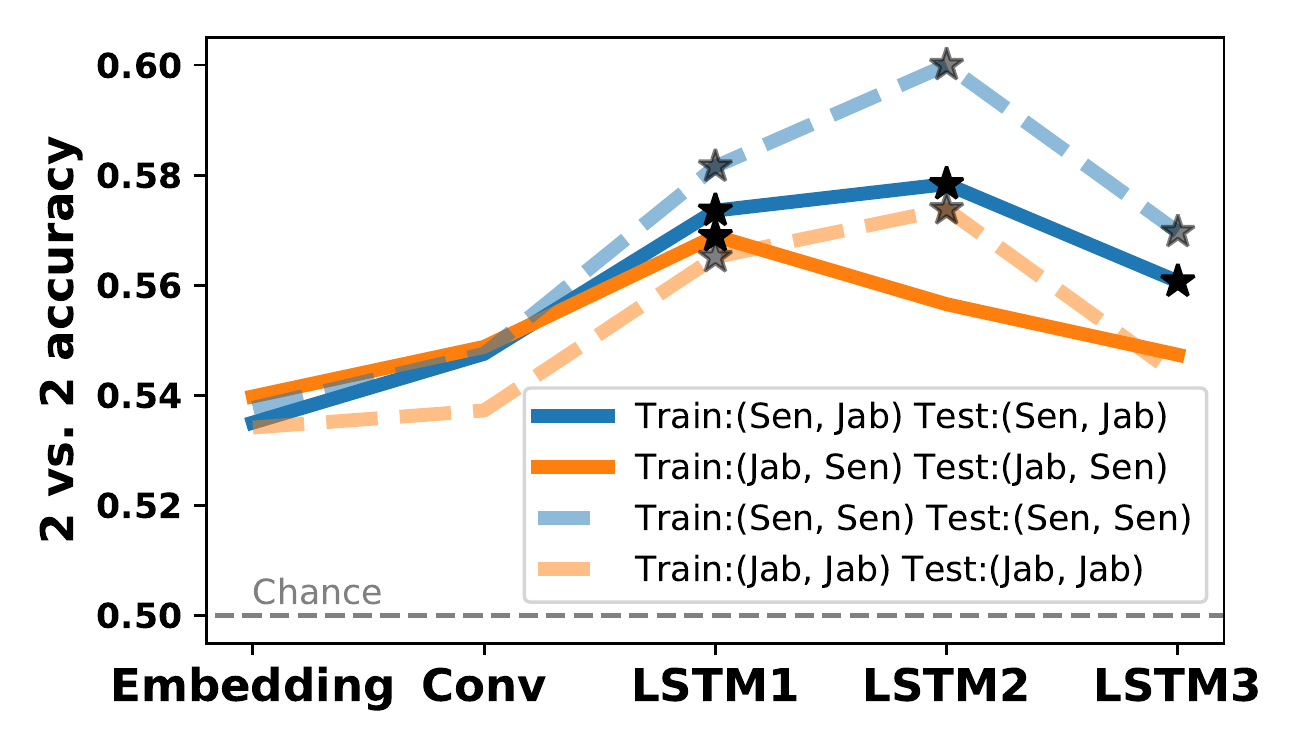}  
  \caption{Analysis 2 (Swap $g(S)$ vectors): Solid lines show the \tvt accuracy of the decoding model that uses the \textit{Sentences} EEG signals to predict the  $g(S)$ vectors from the \textit{Jabberwocky} stimuli, and vice versa.}
  \label{fig:exp2}
\end{subfigure}%
\hspace{0.1 in}
\begin{subfigure}{.48\textwidth}
  \centering
  \includegraphics[width=\linewidth]{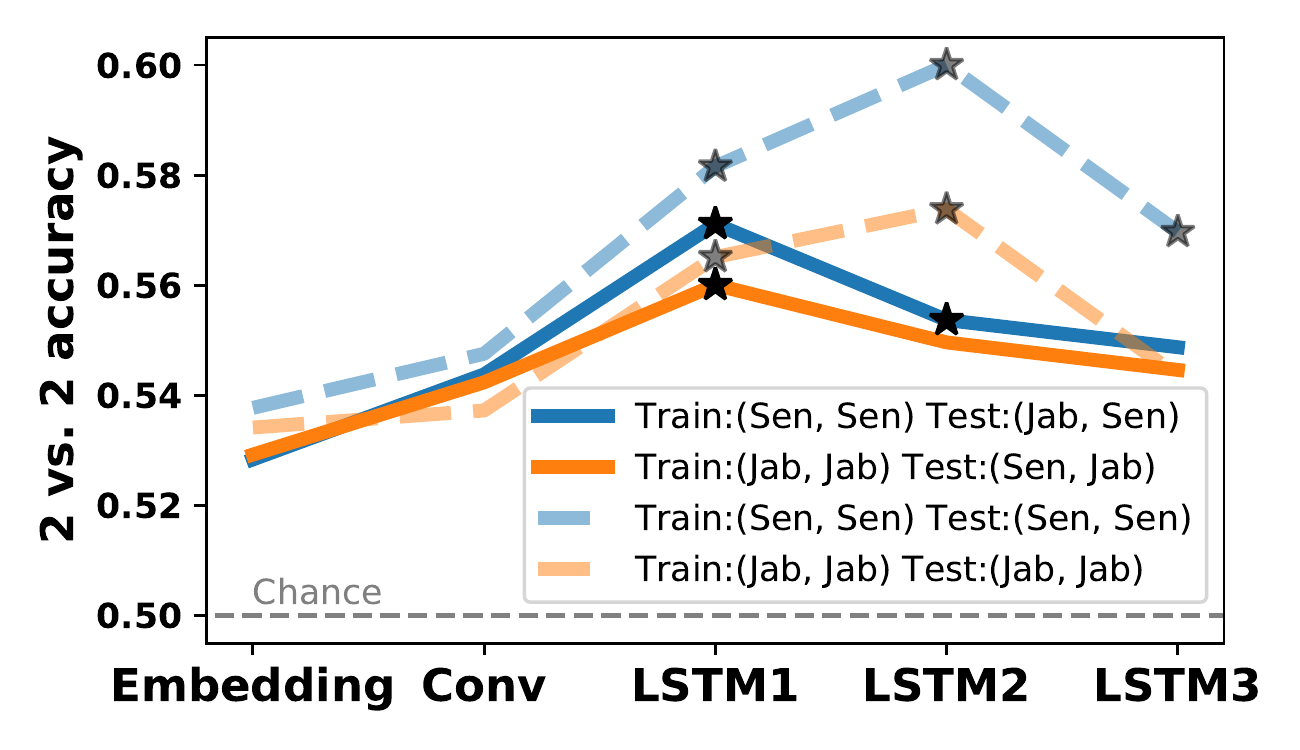} 
\caption{Analysis 3 (Swap $R$ at test time): Solid lines show the \tvt accuracy of the decoding model trained with EEG data and LSTM representations from the same condition, but tested with EEG data from the other condition. }
\label{fig:exp3}
\end{subfigure}
\caption{Results from Analysis 2 and 3. Analysis 1 results appear as dashed lines.  The $x$-axis denotes LSTM representation ($g(S)$).
Legend denotes EEG/LSTM representations used for train/test: (EEG condition, LSTM condition). ``\textit{Sen}'': \textit{Sentence}, ``\textit{Jab}'': \jab, ``\textit{WL}'': \textit{Word-list}. {\bf $\star$}: above chance ($p < 0.05$, FDR corrected).
}
\label{fig:test}
\end{figure*}

In Fig. \ref{fig:exp2} we see that the EEG signals from the \textit{Sentence} condition can be used to predict the \jab LSTM representations (case 1). The accuracies for LSTM1-3 are $0.573$, $0.578$, and $0.560$ which are all above chance ($p < 0.05$, FDR corrected).
For the most part, the accuracies for case 1 are lower than the results from case 1 in Analysis 1 (EEG/LSTM representations from the \sen condition), and we find there is a significant difference in the performance of LSTM2 ($p =0.0006$). This is consistent with the hypothesis that \jab LSTM representations contain only syntactic information.
Interestingly, the \tvt accuracy when using \sen EEG and \jab LSTM representations is higher than Analysis 1, where \jab EEG was paired with \jab LSTM representations.  This is evidence that the syntactic information encoded in the \sen EEG signals may be less noisy. 


In case 2, when we use the \jab EEG to predict the \sen LSTM representations, only the first LSTM layer shows above chance accuracy ($0.568$, $p < 0.05$, FDR corrected).
This implies that the EEG signals from the \textit{Jabberwocky} condition are not significantly correlated with the syntactic information in LSTM2 and LSTM3 vectors derived from \sen stimuli. However, LSTM1 seems to encode syntactic information that is exchangeable. 

Though we did not explicitly test the correlation of the LSTM vectors for the \sen and \jab conditions, Analysis 2 provides evidence that the two may encode correlated syntactic information.  In addition, recall that the LSTM fed \jab can predict the next word of the corresponding \sen stimuli with perplexity close to that of an LSTM fed \sen stimuli.  That predictability is another piece of evidence that the representations share information that could be leveraged in across the two decoding tasks.


\subsection{Swap $R$ at \emph{test} time only (Analysis 3)}

This analysis tests if a trained decoding model can generalize to EEG data from the other condition.  For example, can a model trained with EEG signals and LSTM representations both from the \textit{Sentence} condition still predict the \sen vectors when tested on EEG from the \jab condition? This is Analysis 3 in Table \ref{t:exp1}.
If the \emph{pattern} leveraged to predict LSTM representations is similar across the two conditions, the \tvt accuracy will remain above chance.  

In Fig. \ref{fig:exp3}, for case 1 (train on \sen EEG, test on \jab EEG), the accuracies of the LSTM1 ($0.571$) and LSTM2 ($0.553$) are statistically above chance ($p < 0.05$, FDR corrected). 
Thus, the model trained using \sen EEG can predict \sen vectors from the corresponding \textit{Jabberwocky} EEG.  This implies that the brain's representation for the syntax in both the \sen and \jab conditions takes a similar form, at least with respect to the syntactic information represented in LSTM1 and LSTM2.  However, the performance of LSTM2 here is significantly lower than the performance of LSTM2 in case 1 of Analyses 1 and 2 ($p=0.0001, p=0.0005$ respectively).  In fact, the performance for LSTM2 has dropped by a very large margin compared to Analysis 1, presumably because the semantic information leveraged in Analysis 1 is not available in the \jab EEG.  

For case 2,  (trained on \textit{Jabberwocky} EEG/LSTM representations, but tested on \sen EEG), only LSTM1 can be predicted with above chance \tvt accuracy ($0.560$ with $p=0.001$).  So, as we saw in case 1, the LSTM1 model does generalize to EEG from the other condition.  But, the same cannot be said for LSTM2, which is not significantly above chance in this case.  That LSTM2 generalizes in one direction (case 1) but not the other (case 2) implies that the \jab EEG data is noisier, leading to a less robust model.



\section{Discussion}

Considering the results as a whole, several points become clear.  There is a relationship between the semantic and/or syntactic information as represented by the brain and by LSTM representations, at least for the \sen and \jab conditions.  The probing results are quite consistent with the results of Analyses 1-3: LSTM1 has a strong signal for syntax, LSTM2 has syntax and semantics, and LSTM3 has some syntax and/or semantic signal, but the signal is weaker than for LSTM1-2.

LSTM1 shows only minor changes in performance in Analysis 2 and 3.  So the syntactic information encoded in this layer is fairly consistent for stimuli from both the \sen and \jab conditions, and it correlates well to either EEG data source. There is likely not much semantic information to leverage here, as the performance of models trained on \sen EEG change by only a small amount in Analysis 2 and 3. 

In Analysis 2 we saw similar drops in LSTM2 performance for both \sen and \jab conditions.  The drop in performance using the \sen EEG could be attributed to the lack of semantic information in the \jab LSTM representations.  However, we see a similar size drop in performance for the \jab condition, which implies that there is a mismatch even in the syntactic information available in LSTM2 for the two conditions.  In Analysis 3, when we swap the test data, the pattern learned to predict LSTM2 in the \sen condition (leveraging semantics and syntax) is not as effective when tested on \jab data.

The performance of LSTM3 is harder to explain, possibly because it has weaker semantic/syntactic signal (as evidenced by the probing tasks).  There is a small performance hit when training on \sen EEG data in Analysis 2, but a very large drop in Analysis 3.  This pattern could result if LSTM3's representations of syntax are similar for \sen and \jab stimuli, but the brain showed differing representations for the syntactic information in the two conditions.  Then, it is possible that only the \sen EEG would correlate to the syntactic information in LSTM3.

We wondered if there could be another explanation for our ability to decode in the \jab condition.  One possibility is that the EEG and LSTM layers contain a correlate of the position in a sentence (1st word, 2nd word, etc.), and our models are using that information to decode (7/8 \tvt tests will use words at different positions).  
To test for this possibility, we trained a classifier to predict the ordering of two random words selected from a sentence, as suggested by \citet{adi2016fine}. The input to the classifier is the LSTM representation of the two words at their positions in a sentence, and the output is a binary decision for which of the two words appears sooner in the sentence.  A model trained using our LSTM and the \sen stimuli produced $80 \%$ accuracy on this task.  Thus, we cannot say unequivocally that our results are not due in some part to positional information. However, our probing results are consistent with there being semantic/syntactic information in the representations, and those results are very consistent with the decoding analysis.  This is strong evidence that our results are not entirely due to positional information. 

We wondered also if the lexical semantics of the \jab stimuli could be leaking into the LSTM vectors, perhaps because the pseudo-words were repaired in the convolution step of the LSTM. Note, however, that lexical semantics are entirely intact in the \wl condition, but the LSTM representations are of no use in that condition.  Morphosyntax and syntax are maintained in the \jab condition, which appears to be enough to drive the correlation between LSTM representations and EEG recordings.  The LSTM may be picking up on bi- and tri-gram signals related to morphosyntax cueing syntactic structure \citep{Martin2016,Martin2020}, but more work is needed to rule out alternative explanations.

Recall that the \sen and \jab stimuli share some orthographic/phonological information.  Could our \jab results, and the results of Analysis 2 (swap $g(s)$), be due only to the EEG encoding phonological or orthographic information?  If our models were able to leverage such information, we would expect to see comparable decoding results in Analysis 1 and the \wl condition, where the stimuli are perfect orthographic matches to the EEG. However, that analysis did not produce significantly above-chance accuracy.  Furthermore, if the information leveraged  in Analysis 2 was at the character-level, we would expect to see significantly above-chance accuracy in the character embedding or convolutional layers.  However, it is not until the first LSTM layer (where contextual information is first incorporated) that any decoding model performs significantly above chance in any condition.  This is evidence that the information being leveraged is \emph{not simply phonological or orthographic}.

Our stimuli are composed of two conjoined sentences.  
How much composition have Dutch listeners done by the time when they get to the conjunction word ``en?'' How does the processing differ between the first vs the second of the conjoined sentences?
Previous work on the brain's processing of syntactic structures and coordinate clauses proposed an ``active structure maintenance model'', where neural activity increases as a function of syntactic complexity~\citet{pallier2011cortical,lau2018linguistic}. They found that neural activity in certain left-hemispheric regions indeed increased when more constituents had to be integrated, for both sentences and jabberwocky stimuli. It may be that the second  coordinate constituent in our stimuli sentences elicit stronger neural activity than the first, but more analysis would be required to verify this.

\section{Related Work}

The first example of mapping brain responses onto corpus-derived  representations appeared in \newcite{mitchell2008predicting}. This study encoded word meaning into vectors of word co-occurrence features. The authors showed that a trained linear regression model could predict fMRI activation in response to single concrete noun stimuli.  From there, decoding models were shown to work with dependency-parse-based representations \citep{murphy2012selecting} and with concept-relation-features extracted from topic models \citep{pereira2013using}. 
\newcite{anderson2017visually} demonstrated that decoding models can learn the pattern of the brain's response to abstract concepts/nouns. 

Some of the first examples of decoding language \emph{in context} were from \citet{wehbe2014simultaneously} and ~\citet{de2017hierarchical}.  The first models used a combination of (non-contextual) corpus-derived, acoustically-derived and/or hand-coded representations. Several groups then began to experiment with encoding models based on \emph{contextual} language representations, like those in recurrent neural network (RNN) language models \cite{wehbe2014aligning,jain2018incorporating,toneva2019interpreting}. These models showed that vectors incorporating contextual information could be decoded from brain imaging data, and contextual models actually outperformed non-contextual word vectors.  We confirmed those findings here. 

Though there are fewer decoding models trained on EEG, there are a few recent examples.  \citet{hale2018finding} showed that the operations performed by an RNN-grammar trained to parse sentences correlated to EEG collected while people listened to a story. \citet{schwartz2019understanding} found connections between bi-LSTM representations and the ERPs (event related potentials) more classically used to study language in the brain.  Our work adds to the new body of work showing that EEG can be a powerful data source in this space.

\section{Conclusion and future work}

In this study, we explored the correlation of a character-level LSTM with the brain's response for two kinds of out-of-distribution language. The \jab condition used pseudo-word translations of the \sen stimuli (ablate semantics, preserve syntax).  The \wl stimuli was a pseudo-random re-ordering of the words in each of the \sen stimuli (ablate syntax, preserve semantics).  We ran a character-based LSTM to create contextual embeddings for the stimuli of each condition.  Our linear-regression decoding models were trained to predict the various LSTM representations from the EEG signals.

Our results showed that the LSTM layers of this character-based LSTM do in fact correlate with EEG signals in both the \sen and \jab conditions, but not in the \wl condition.  By training models with various alterations to the data, we were able to determine which LSTM representations carry semantic and syntactic information.  We verified those results using a probing task on our Dutch LSTM, as well as an identical model trained on English.

There are multiple avenues for future work.  For example, Dutch has a fairly transparent phoneme-grapheme correspondence; would our results still hold for a language with deeper orthography?  We were surprised to find that some LSTM representations resembled the \jab EEG signals. Are there other examples of out-of-distribution language where this relationship holds?  And, perhaps more interestingly, where it does not hold?  Finding ways in which the brain's representations differ from an LSTM could help us to build models closer to the true nature of human language processing.


\section*{Acknowledgments}
AF and MW are supported by the  Natural Sciences and Engineering Research Council of Canada (NSERC) Discovery Grants, and hold Canada CIFAR (Canadian Institute for Advanced Research) AI Chairs.  The computational work was supported in part by infrastructure made available by WestGrid and Compute Canada. AEM was supported by the Max Planck Research Group ``Language and Computation in Neural Systems'' and by the Netherlands Organization for Scientific Research (grant 016.Vidi.188.029).


\bibliographystyle{acl_natbib}
\bibliography{emnlp2020}

\newpage
\appendix

\section{Supplemental Material: Probing task performance}
\label{app:A}
Table \ref{t:probDef} describes the probing tasks in English from \citet{conneau2018you} and in Dutch from \newcite{eichler-etal-2019-linspector}. Table \ref{t:resultProbEng} shows probing task accuracy for both English and Dutch datasets, as measured with the character-based LSTMs proposed by \citet{kim2016character}.  The English model is trained on the Penn Treebank \cite{marcus1993building}, the Dutch on Dutch Wikipedia.

\begin{table*}
  \begin{center}
    \caption{Description of the probing tasks. ``En'' shows the English datasets and ``Du'' shows the Dutch datasets.}
    \label{t:probDef}
    \begin{tabular}{|c|l|l|c|} 
    \hline
      \textbf{Type} & \textbf{Name} & \textbf{Description}& \textbf{Data}\\
      \hline
      \rowcolor{Gray}
        &\text{Tense} &\textit{Tense of the main-clause verb (present/ past)} & \textit{En/Du}\\ \cline{2-4}
        \rowcolor{Gray}
        &\text{Subject number} &\textit{Number of the subjects of the main clause} & \textit{En} \\ \cline{2-4}
        \rowcolor{Gray}
        &\text{Object number} &\textit{Number of the direct objects of the main clause} & \textit{En} \\ \cline{2-4}
        \rowcolor{Gray}
        \multirow{-4}{*}{\textbf{Semantic}} &\text{Coordination inversion} &\textit{Indicate if a sentence is intact or modified
        } & \textit{En} \\ \cline{2-4}
      \hline
        \rowcolor{LightCyan}
        &\text{Bigram shift} &\textit{Indicate having legal word orders } & \textit{En}\\ \cline{2-4}
        \rowcolor{LightCyan}
        &\text{Tree depth} &\textit{Depth of the hierarchical structure of sentences} & \textit{En} \\ \cline{2-4}
        \rowcolor{LightCyan}
         &\text{Top constituent} &\textit{Indicate top constituent sequence of sentences} & \textit{En}\\ \cline{2-4}
         \rowcolor{LightCyan}
        &\text{Number} &\textit{Indicate singularity and plurality of nouns/adjectives/verbs } & \textit{Du}\\ \cline{2-4}
        \rowcolor{LightCyan}
         \multirow{-5}{*}{\textbf{Syntactic}}&\text{Part of Speech} &\textit{Indicate the part of speech of a specific word} & \textit{Du}\\ \cline{2-4}
      \hline
    \end{tabular}
  \end{center}
\end{table*}

\begin{table*}
  \begin{center}
    \caption{Probing task accuracies. Each row shows the accuracies of a specific probing task described in Table~\ref{t:probDef}. Columns correspond to the LSTM representation: ``\textit{Embedding}'': Embedding layer, ``\textit{Conv}'': concatenation of Convolutional layers, ``\textit{LSTM1-3}'': an LSTM layers. ``\textit{Tense/En}'' and ``\textit{Tense/Du}'' denote the English and Dutch probing task for \textit{Tense}, respectively.}
    \label{t:resultProbEng}
    \begin{tabular}{|l|c|c|c|c|c|} 
    \hline
      \textbf{Layers \#} & \textbf{Embedding} & \textbf{Conv} &\textbf{LSTM1} & \textbf{LSTM2} & \textbf{LSTM3} \\
      \hline
      \rowcolor{Gray}
      \textbf{Tense/En}& 43.2 & 53.2 & 63.2& 70.7 & 63.9 \\
      \hline
      \rowcolor{Gray}
      \textbf{Tense/Du}& 46.4	& 55.1 & 66.7 & 72.7 & 62.7 \\
      \hline
      \rowcolor{Gray}
      \textbf{Subject number}& 38.8	& 53.5 & 65.5& 72.1 & 64.3 \\ 
      \hline
      \rowcolor{Gray}
      \textbf{Object number}& 39.5 & 52.1& 66.8& 71.7 & 65.8 \\
      \hline
      \rowcolor{Gray}
      \textbf{Coord. Inv.}& 40.5 & 46.6 & 58.7& 66.1 & 61.3 \\ 
      \hline
      \rowcolor{LightCyan}
      \textbf{Bigram shift}& 43.1 & 53.1 &70.8& 69.4 & 58 \\
      \hline
      \rowcolor{LightCyan}
      \textbf{Tree depth}& 39.3 &	45.6 & 56.3& 58.6 & 54.3 \\
      \hline
      \rowcolor{LightCyan}
      \textbf{Top constituent} & 38.5	& 53.8 &75.5& 76.1 & 64.1 \\
      \hline
      \rowcolor{LightCyan}
      \textbf{Number}& 52.3 & 58.6 & 78.2 & 81.9 & 67.3 \\
      \hline
      \rowcolor{LightCyan}
      \textbf{Part of Speech} & 39.6 & 53.7 & 69.8 & 76.1 & 60.3 \\
      \hline
    \end{tabular}
  \end{center}
\end{table*}

\section{ Supplemental Material: Measuring model accuracy by mean-squared-error}
\label{app:B}

\begin{figure}
\centering
\includegraphics[width=\linewidth]{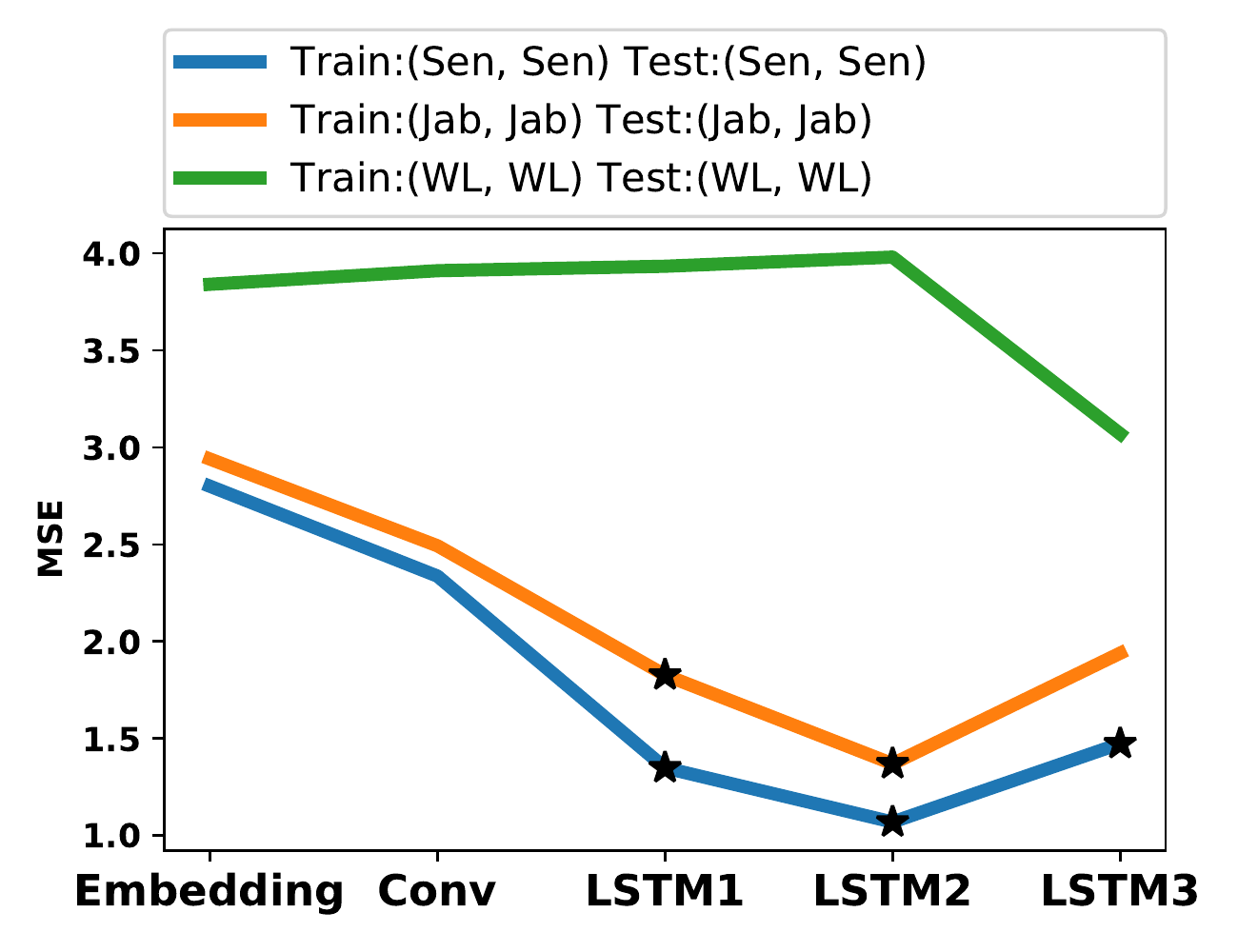}  
\caption{Analysis 1 (Test for semantic and syntactic information): MSE for  $g(S)$/EEG from the same condition.   The $x$-axis denotes LSTM representation ($g(S)$).
Legend denotes EEG/LSTM representations used for train/test: (EEG condition, LSTM condition). ``\textit{Sen}'': \textit{Sentence}, ``\textit{Jab}'': \jab, ``\textit{WL}'': \textit{Word-list}. {\bf $\star$}: below chance ($p < 0.05$, FDR corrected).
}
\label{fig:exp1_mse}
\end{figure}

\begin{figure*}
\centering
\begin{subfigure}{.48\textwidth}
  \centering
  \includegraphics[width=\linewidth]{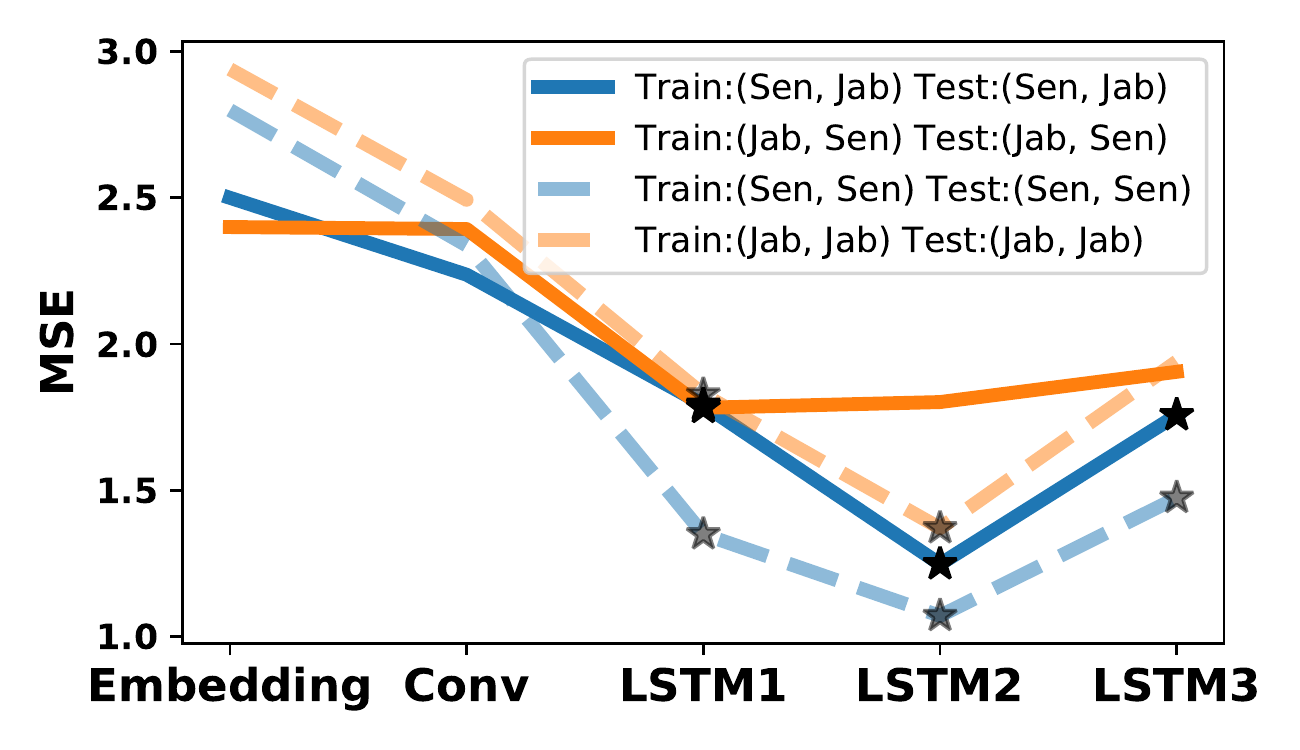}  
  \caption{Analysis 2 (Swap $g(S)$ vectors): Solid lines show the MSE of the decoding model that uses the \textit{Sentences} EEG signals to predict the  $g(S)$ vectors from the \textit{Jabberwocky} stimuli, and vice versa.}
  \label{fig:exp2_mse}
\end{subfigure}%
\hspace{0.1 in}
\begin{subfigure}{.48\textwidth}
  \centering
  \includegraphics[width=\linewidth]{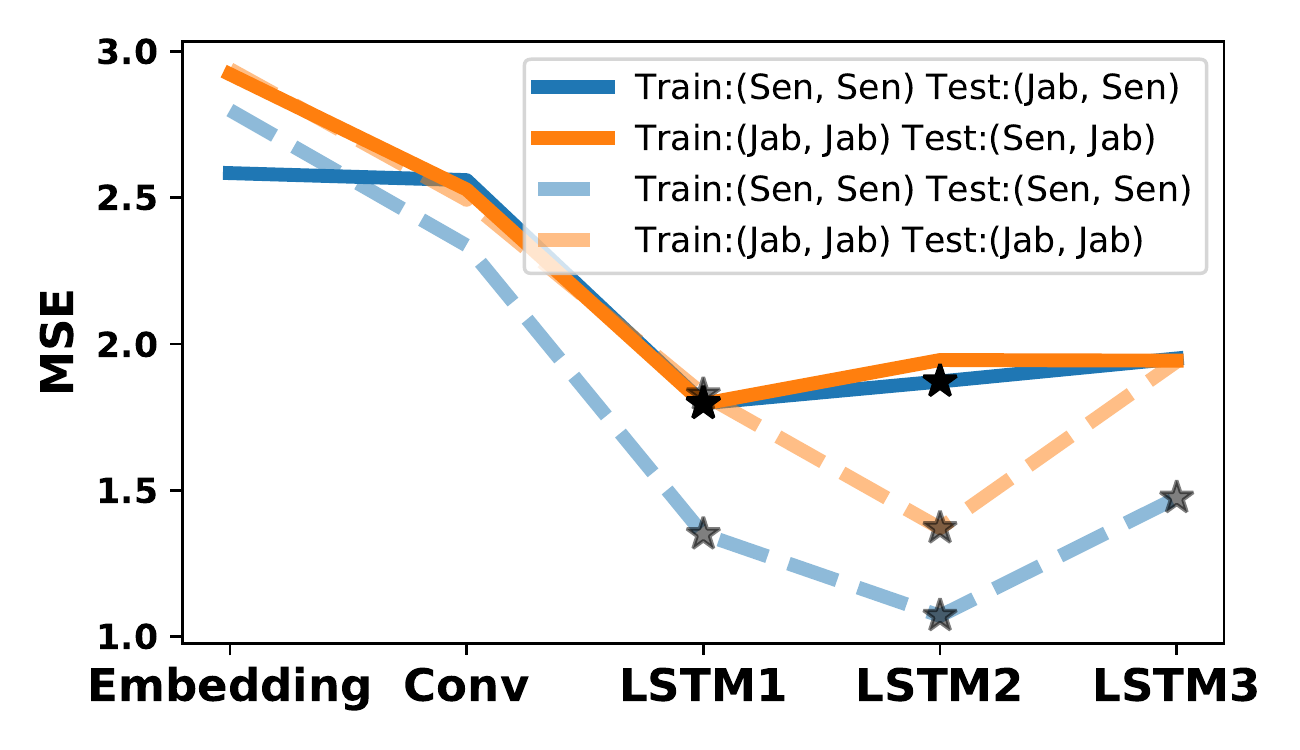} 
\caption{Analysis 3 (Swap $R$ at test time): Solid lines show the MSE of the decoding model trained with EEG data and LSTM representations from the same condition, but tested with EEG data from the other condition. }
\label{fig:exp3_mse}
\end{subfigure}
\caption{MSE results from Analysis 2 and 3. Analysis 1 results appear as dashed lines.  The $x$-axis denotes LSTM representation ($g(S)$).
Legend denotes EEG/LSTM representations used for train/test: (EEG condition, LSTM condition). ``\textit{Sen}'': \textit{Sentence}, ``\textit{Jab}'': \jab, ``\textit{WL}'': \textit{Word-list}. {\bf $\star$}: below chance ($p < 0.05$, FDR corrected).
}
\label{fig:test_mse}
\end{figure*}

In addition to 2 vs. 2 accuracy, we also used mean-squared-error (MSE) to assess the performance of the decoding model. Figs.~\ref{fig:exp1_mse} and \ref{fig:test_mse} show the results of MSE for analyses 1-3.

\end{document}